\documentclass[AMA,STIX1COL]{WileyNJD-v2}

\articletype{RESEARCH ARTICLE}%

\received{}
\revised{}
\accepted{}

\usepackage{bm}

\begin{document}
	
	\title{Tensor-based Intrinsic Subspace Representation Learning for Multi-view Clustering}
	
	\author[1]{Qinghai Zheng}
	\author[1]{Yu Zhang}
	\author[1]{Jihua Zhu*}
	\author[1]{Zhongyu Li}
	\author[1]{Haoyu Tang}
	\author[2]{Shuangxun Ma}
	
	\authormark{Qinghai Zheng \textsc{et al}}
	
	\address[1]{\orgdiv{Lab of Vision Computing and Machine Learning, School of Software Engineering}, \orgname{Xi'an Jiaotong University}, \orgaddress{\state{Xi’an 710049}, \country{China}}}
	
	\address[2]{School of Software Engineering, Xi'an Jiaotong University, Xi’an 710049, China}
	
	\corres{*Jihua Zhu. \email{zhujh@xjtu.edu.cn}}
	
	\abstract[Abstract]{As a hot research topic, many multi-view clustering approaches are proposed over the past few years. Nevertheless, most existing algorithms merely take the consensus information among different views into consideration for clustering. Actually, it may hinder the multi-view clustering performance in real-life applications, since different views usually contain diverse statistic properties. To address this problem, we propose a novel Tensor-based Intrinsic Subspace Representation Learning (TISRL) for multi-view clustering in this paper. Concretely, the rank preserving decomposition is proposed firstly to effectively deal with the diverse statistic information contained in different views. Then, to achieve the intrinsic subspace representation, the tensor-singular value decomposition based low-rank tensor constraint is also utilized in our method. It can be seen that specific information contained in different views is fully investigated by the rank preserving decomposition, and the high-order correlations of multi-view data are also mined by the low-rank tensor constraint. The objective function can be optimized by an augmented Lagrangian multiplier based alternating direction minimization algorithm. Experimental results on nine common used real-world multi-view datasets illustrate the superiority of TISRL.}
	
	
	\maketitle

	\section{Introduction}
	
	Clustering is a fundamental and interesting research topic, the goal of which is to group unlabeled data samples into corresponding categories according to their underlying similarities or relationships. Many clustering methods have been proposed, such as k-means clustering \cite{jain2010kmeans}, spectral clustering \cite{von2007tutorial}, fuzzy clustering \cite{wang2015large}, and subspace clustering \cite{vidal2011subspacec}. However, these methods are used to deal with single view data rather than multi-view data \cite{xu2013survey}, which describe subjects from multiple domains or various types of features. With the development of technology and measurement, data with multiple views are more commonly seen in real-life applications. For example, a video can be expressed by visual signal and audio signal simultaneously, an image can be depicted by the LBP feature (local binary pattern) and SIFT feature (scale-invariant feature transform) comprehensively, genes can be described by their expression levels and their somatic mutation in different cellular environments. Obviously, clustering algorithms designed for single view data are no longer suitable for dealing with the multi-view data \cite{zheng2019feature,xue2019deep,gLMSC2020PAMI,huang2014co}. To excavate more information contained in multi-view data, many multi-view clustering methods, e.g., multi-view subspace clustering and multi-view spectral clustering, are proposed to enhance the clustering performance in recent years \cite{xu2013survey,gLMSC2020PAMI,chao2017survey,tang2018learning,zheng2020constrained}. In this paper, we focus on multi-view subspace clustering.
	
	Multi-view subspace clustering usually learns the subspace representation of different views and gets clustering results by employing the standard spectral clustering \cite{zheng2019feature,gLMSC2020PAMI,chao2017survey,zheng2020constrained,di2019lightweight,yang2018surveymultiview,zhou2019dual}. The major difference of tese existing methods depends on the means of the investigating approaches of subspace representations. For example, the method proposed in \cite{zheng2019feature} utilizes concatenated features to learn a desired subspace representation to get clustering results. The algorithm proposed in \cite{gLMSC2020PAMI} learns a latent representation to achieve the subspace representation learning, in which the complementary information can be effectively investigated. The method proposed in \cite{zhou2019dual} obtains a subspace representation by pursuing the dual shared-specific subspace representation to explore the correlations and common-shared information of multi-view data. Recently, some tensor-based approaches are also proposed by exploring the high-order correlations for multi-view subspace clustering \cite{zhang2015low,xie2018unifying,2020Tensorized}. For instance, the method proposed in \cite{zhang2015low} learns the subspace representation by leveraging a generalized tensor nuclear norm based low rank tensor constraint \cite{liu2012tensor} to mine the high order correlations of different views, and the approach proposed in \cite{xie2018unifying} uses the tensor-Singular Value Decomposition (t-SVD) based low-rank tensor constraint \cite{kilmer2013third} for learning. The difference between works proposed in \cite{zhang2015low} and \cite{xie2018unifying} is the tensor norm used in their works. 
	
	Although significant improvements have been attained, it can be observed that most existing multi-view subspace clustering methods, including the tensor-based approaches, apply some specific constraints, e.g., the low-rank tensor constraint, on the subspace representation matrices straightforward, and consequently, ignore the specific information contained in multi-view data. In practice, different views usually have diverse even incompatible statistic properties, these above-mentioned approaches may not achieve good clustering results in real-life applications.
	
	\begin{figure}[!t]
		\centering
		\includegraphics[width=1\columnwidth]{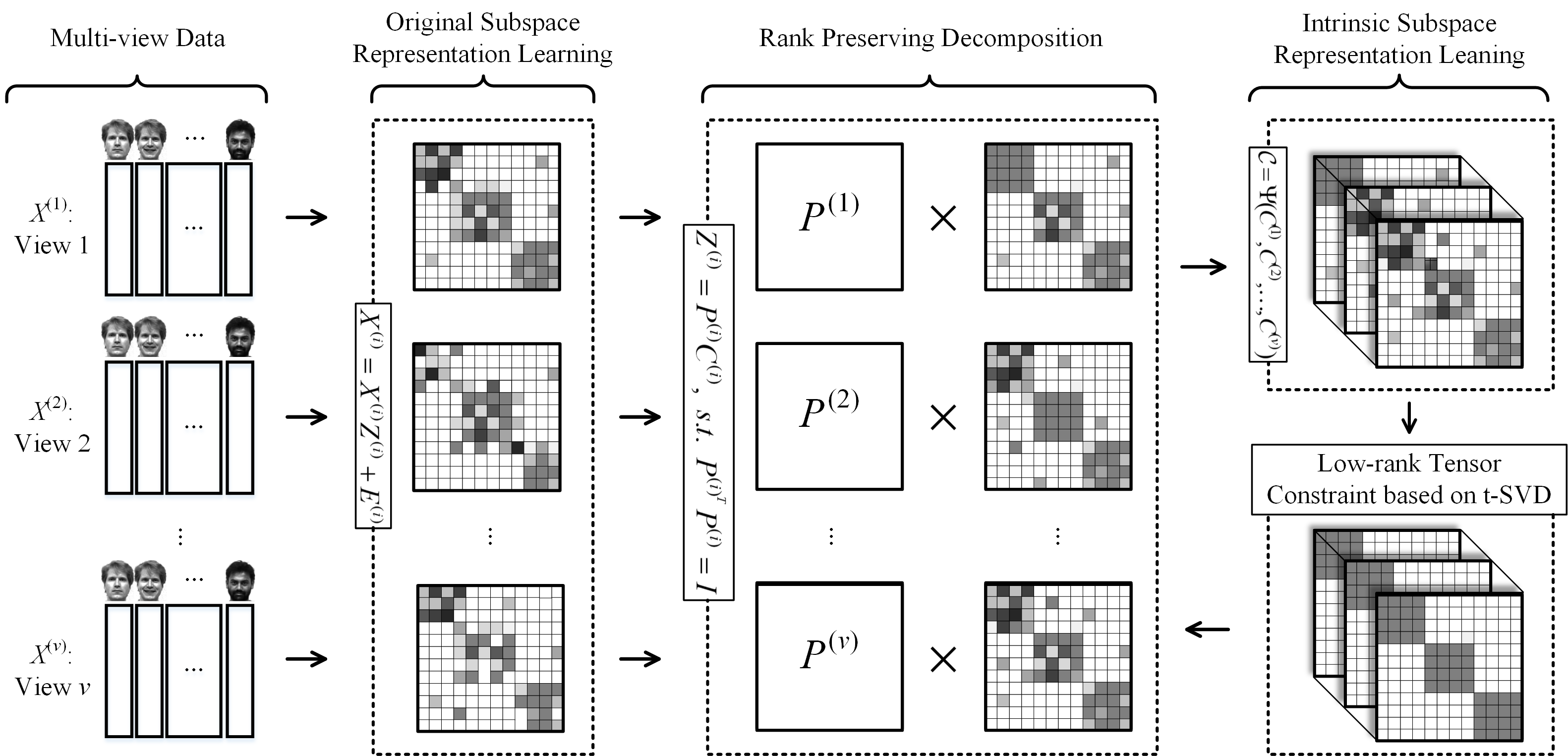}
		\caption{The whole framework of the proposed TISRL. Given multi-view data, i.e., $\{ {\bm{X}^{(i)}}\} _{i = 1}^v$, the original subspace representations, i.e., $\{ {\bm{Z}^{(i)}}\} _{i = 1}^v$, are learned and decomposed into orthogonal matrices, i.e., $\{ {\bm{P}^{(i)}}\} _{i = 1}^v$, and rank preserving matrices, i.e., $\{ {\bm{C}^{(i)}}\} _{i = 1}^v$. Then, the t-SVD-based low-rank tensor constraint is imposed and the intrinsic subspace representation can be learned elegantly. It can be observed that the specific information of multi-view data can be considered in $\{ {\bm{P}^{(i)}}\} _{i = 1}^v$, and clustering results can be obtained by performing the spectral clustering on the learned intrinsic subspace representation.}
		\label{TISRL_Flowchart}
	\end{figure}
	
	To deal with aforementioned problem, a novel Tensor-based Intrinsic Subspace Representation Learning, termed TISRL, is proposed in this paper. The whole framework is depicted in Fig.~\ref{TISRL_Flowchart}. To be specific, the original subspace representations are learned and decomposed into orthogonal matrices and rank preserving matrices firstly. Different from most existing methods, which impose some specific constraints on the original subspace representation matrices directly \cite{zhang2015low,xie2018unifying}, the proposed method performs the rank-preserving decomposition on original subspace representation matrices, and then employs the t-SVD based low-rank tensor constraint to learn the intrinsic representation matrix elegantly. The rank-preserving decomposition introduced and applied in TISRL has several appealing characteristics: a) it considers and explores the diverse statistical information brought by multiple views ; b) no extra parameter will be added by the strategy of introducing the rank-preserving decomposition.
	To optimize the objective function effectively, an alternating direction minimization algorithm based the Augmented Lagrangian Multiplier (ALM) \cite{lin2011linearized} is also proposed in this paper. The main contributions of TISRL can be summarized and provided as follows:
	
	\begin{enumerate}
		\item We introduces the rank-preserving decomposition, which is applied on the original subspace representations. The specific information of multi-view data can be taken into consideration by the rank-preserving decomposition.
		\item We adopt the tensor-singular value decomposition based low-rank tensor constraint on the learned new matrices to learn the desired intrinsic subspace representation. The high-order correlations of multiple views can be fully investigated for clustering.
		\item We design an alternating direction minimization algorithm based on ALM for the optimization of TISRL's objective function. We also carried out experiments on nine real-life datasets to illustrate the advantages of TISRL approach.
	\end{enumerate}
	
	\section{Related Work}
	As an important and interesting problem, extensively related works of multi-view clustering have been conducted and proposed in recent years \cite{xu2013survey,yang2018surveymultiview,zhang2019multitask}. The basic motivation of multi-view clustering is to enhance clustering results with the abundant information, such as consensus and complementary information, contained in multiple views \cite{chao2017survey}. In a big picture, most existing multi-view clustering methods can be categorized into three types roughly \cite{zhang2015low,xie2018unifying,xia2014robust,wang2019gmc,peng2019comic,zhang2018binary,yukernel2020}, including the graph based methods, i.e., multi-view spectral clustering, the subspace representation based methods, i.e., multi-view subspace clustering, and other multi-view clustering methods.
	
	In the graph based methods \cite{xia2014robust,wang2019gmc,yukernel2020,zhan2017graph,wu2019essential,zhou2019incremental,wang2019study,kang2020multi}, graph information of multiple views is investigated to promote clustering results. For example, the method proposed in \cite{xia2014robust} uses the Markov chain clustering algorithm and constructs similarity matrices of different views firstly and then finds a latent transition probability matrix of all views via the low rank and sparse decomposition. The method proposed in \cite{zhan2017graph} considers graph relationships among multiple views and excavates the consensus property contained in different views to get clustering results. The method proposed in \cite{wang2019gmc} works on the underlying similarity matrix for clustering by mining the graph information of all views and also taking the weights of different views into consideration. The method proposed in \cite{wu2019essential} utilizes a t-SVD-based essential tensor learning strategy to dig the desired low-rank properties and improves clustering results by leveraging the principle information of different views.
	
	As for the second type, i.e., multi-view subspace clustering \cite{tang2018learning,cao2015diversity,zhang2015low,xie2018unifying,gLMSC2020PAMI,zheng2019feature,li2020robust,brbic2018multi,zhang2020one}, different approaches are mainly discriminated on the way of subspace representation learning, and variety types of constraints are imposed on the original subspace representations for learning. For example, the method proposed in \cite{cao2015diversity} imposes the HSIC-based diversity constraint on subspace representations of different views to explore the complementary information. The method proposed in \cite{gLMSC2020PAMI} seeks a latent feature representation and learns the corresponding subspace representation at the same time to get clustering results based on the learned subspace representation. The method proposed in \cite{zheng2019feature} conducts subspace clustering on the concatenated views to obtain results by introducing the cluster-specific corruptions brought by different views. Some low-rank tensor based approaches are also conducted in \cite{zhang2015low,xie2018unifying}. Low-Rank Tensor Constrained Multi-view Subspace Clustering (LT-MSC) proposed in \cite{zhang2015low,zhang2020tensorized} is an algorithm, which uses the low-rank tensor constraint to achieve clustering results, and it employs the unfolding based tensor norm \cite{liu2012tensor}. The t-SVD based Multi-view Subspace Clustering (t-SVD-MSC) introduces a t-SVD-based tensor norm \cite{kilmer2013third} to explore the complementary information and view correlations in the high-order levels.  
	
	Furthermore, some algorithms belonging to other categories are also proposed in recent years \cite{peng2019comic,wang2015deep,zhao2017multiDMF,chen2020robust}. For example, the work conducted in \cite{zhao2017multiDMF} performs the deep matrix factorization on multiple feature representations to learn the compact multi-view representation and achieves clustering results by using spectral clustering algorithm. The method proposed in \cite{peng2019comic} draws lessons from the continuous clustering and attains multi-view clustering results by mining consistency among different views both on the geometric and the cluster assignment. 
	
	\section{The Proposed Approach}
	In this section, for the clear expression, the frequently used notations and the tensor's preliminaries are summarized here firstly, then the proposed TISRL and its optimization algorithm are introduced in detail.
	
	\begin{table}[!t]
		\caption{Notations used in this paper.}
		\label{table_symbols}
		\centering
		\resizebox{0.53\textwidth}{!}{
			\begin{tabular}{l|l|l|l}
				\toprule
				$a$ & A scale & $\bm{a}$ & A vector \\
				\midrule
				$v$ & Number of views & $n$ & Number of samples \\
				\midrule
				$\bm{A}$ & A mateix  & $\bm{{\cal A}}$ & A tensor \\
				\midrule
				rank($\bm{A}$) & Rank of $\bm{A}$ & ${\left\| \bm{A} \right\|_ * }$ & Nuclear norm of $\bm{A}$\\
				\midrule
				${\left\| \bm{A} \right\|_{2,1}}$ & $l_{2,1}$-norm of $\bm{A}$ & ${\left\| \bm{A} \right\|_F}$ & Frobenius norm of $\bm{A}$\\
				\midrule
				${\bm{{\cal A}}_{ijk}}$ & \multicolumn{3}{l}{The $(i,j,k)$-th element of $\bm{{\cal A}}$} \\
				\midrule
				${\bm{{\cal A}}{(:,i,j)}}$ & \multicolumn{3}{l}{Model-1 fiber of $\bm{{\cal A}}$} \\
				\midrule
				${\bm{{\cal A}}{(i,:,j)}}$ & \multicolumn{3}{l}{Model-2 fiber of $\bm{{\cal A}}$} \\
				\midrule
				${\bm{{\cal A}}{(i,j,:)}}$ & \multicolumn{3}{l}{Model-3 fiber of $\bm{{\cal A}}$} \\
				\midrule
				${\bm{{\cal A}}{(k,:,:)}}$ & \multicolumn{3}{l}{The $k$-th horizontal slice of $\bm{{\cal A}}$} \\
				\midrule
				${\bm{{\cal A}}{(:,k,:)}}$ & \multicolumn{3}{l}{The $k$-th lateral slice of $\bm{{\cal A}}$} \\
				\midrule
				${\bm{{\cal A}}{(:,:,k)}}$ & \multicolumn{3}{l}{The $k$-th frontal slice of $\bm{{\cal A}}$, written as ${\bm{{\cal A}}^{(k)}}$ as well}\\
				\midrule
				${{\bm{{\cal A}}}}^{T}$ & \multicolumn{3}{l}{Transpose of $\bm{{\cal A}}$}\\
				\midrule
				${{\left\|{\bm{{\cal A}}}\right\|}_ {\circledast} }$ & \multicolumn{3}{l}{t-SVD based tensor nuclear norm of $\bm{{\cal A}}$}\\
				\midrule
				${\bm{{\cal A}}_{f}}$ & \multicolumn{3}{l}{$={\rm{fft}({\bm{{\cal A}}},[~],3)}$, Fourier transform along the $3$-rd dimention} \\
				\midrule
				${\left\| {\bm{{\cal A}}} \right\|_F}$ & \multicolumn{3}{l}{$=\sqrt {\sum\nolimits_{i,j,k} {{{\left| {{{\bm{{\cal A}}}_{i,j,k}}} \right|}^2}} }$, Frobenius norm of $\bm{{\cal A}}$}\\
				\bottomrule
		\end{tabular}}
	\end{table}
	
	\subsection{Notations and Preliminaries}
	To be clear, notations frequently used in this paper are given in Table~\ref{table_symbols}. Specifically, the lower case letters, bold lower case letters, bold upper case letters, and bold calligraphy letters are used to indicate the scales, vectors, matrices and tensors, respectively. A 3-order tensor ${\bm{{\cal A}}} \in {{\mathbb{R}}^{{n_1} \times {n_2} \times {n_3}}}$ is considered here. The transpose of $\bm{{\cal A}}$ can be attained by transposing each frontal slice and then reversing the order of transposed frontal slices 2 through $n_3$, i.e., ${{{\bm{{\cal A}}}}^{T}} \in {{\mathbb{R}}^{{n_2} \times {n_1} \times {n_3}}}$ \cite{kilmer2013third}. For the notation of the Fourier transform employed here, it is the same with Matlab command and the corresponding inverse operation can be written as $\bm{{\cal A}} = {\rm{ifft}}({\bm{{\cal A}}_{f}},[~],3)$. Additionally, according to \cite{kilmer2013third}, some important operations of a tensor can be defined as follows. 
	
	The block vectorizing $\rm{unfold}$ and its inverse operator $\rm{fold}$ of a tensor ${\bm{{\cal A}}} \in {{\mathbb{R}}^{{n_1} \times {n_2} \times {n_3}}}$ are as follows:
	\begin{equation}
	{\rm{unfold}}({\bm{{\cal A}}}) = \left[ \begin{array}{l}
	{{\bm{A}}^{(1)}}\\
	{{\bm{A}}^{(2)}}\\
	\vdots \\
	{{\bm{A}}^{({n_3})}}
	\end{array} \right] \in {{\mathbb{R}}^{{n_1}{n_3} \times {n_2}}},
	\end{equation}
	and ${\bm{{\cal A}}}=\rm{fold}(\rm{unfold}({{\bm{{\cal A}}}}))$.
	
	The $\rm{bdiag}$ operation of a tensor ${\bm{{\cal A}}} \in {{\mathbb{R}}^{{n_1} \times {n_2} \times {n_3}}}$ unfolds ${\bm{{\cal A}}}$ to a matrix with the following block-diagonal form:
	\begin{equation}
	\rm{bdiag}({\bm{{\cal A}}}) = \left[ {\begin{array}{*{20}{c}}
		{{{\bm{A}}^{(1)}}}&{}&{}\\
		{}& \ddots &{}\\
		{}&{}&{{{\bm{A}}^{({n_3})}}}
		\end{array}} \right]\in {{\mathbb{R}}^{{n_1}{n_3} \times {n_2}{n_3}}}.
	\end{equation}
	
	The $\rm{bcirc}$ operation of a tensor ${\bm{{\cal A}}} \in {{\mathbb{R}}^{{n_1} \times {n_2} \times {n_3}}}$ is defined as follows:
	\begin{equation}
	\rm{bcirc}({\bm{{\cal A}}}) = \left[ {\begin{array}{*{20}{c}}
		{{{\bm{A}}^{(1)}}}&{{{\bm{A}}^{({n_3})}}}& \cdots &{{{\bm{A}}^{(2)}}}\\
		{{{\bm{A}}^{(2)}}}&{{{\bm{A}}^{(1)}}}& \cdots &{{{\bm{A}}^{(3)}}}\\
		\vdots & \vdots & \ddots & \vdots \\
		{{{\bm{A}}^{({n_3})}}}&{{{\bm{A}}^{({n_3} - 1)}}}& \cdots &{{{\bm{A}}^{(1)}}}
		\end{array}} \right] \in {{\mathbb{R}}^{{n_1}{n_3} \times {n_2}{n_3}}}.
	\end{equation}
	
	An identity tensor ${\bm{{\cal I}}} \in {{\mathbb{R}}^{{n_1} \times {n_1} \times {n_3}}}$ is a tensor, in which the $1$-st frontal slice is an identity matrix $\in {\mathbb{R}}^{{n_1} \times {n_1}}$ and elements on the remain frontal slices are all zeros.
	
	Based on these operations, t-product of two tensors can be defined as follows:
	\begin{equation}
	{\bm{{\cal A}}} * {\bm{{\cal B}}} = {\rm{fold}}({\rm{bcirc}}({\bm{{\cal A}}}){\rm{unfold}}({\bm{{\cal B}}})) \in {{\mathbb{R}}^{{n_1} \times {n_4} \times {n_3}}},
	\end{equation}
	where ${\bm{{\cal A}}} \in {{\mathbb{R}}^{{n_1} \times {n_2} \times {n_3}}}$ and ${\bm{{\cal B}}} \in {{\mathbb{R}}^{{n_2} \times {n_4} \times {n_3}}}$. 
	
	An orthogonal tensor ${\bm{{\cal A}}} \in {{\mathbb{R}}^{{n_1} \times {n_1} \times {n_3}}}$ satisfies the following equation:
	\begin{equation}
	{\bm{{\cal A}}} * {{\bm{{\cal A}}}^T} = {{\bm{{\cal A}}}^T} * {\bm{{\cal A}}} = {\bm{{\cal I}}}.
	\end{equation}
	
	Based on the definition proposed in \cite{kilmer2013third}, the tensor Singular Value Decomposition, i.e., t-SVD, of a tensor ${\bm{{\cal A}}} \in {{\mathbb{R}}^{{n_1} \times {n_2} \times {n_3}}}$ can be formulated as follows:
	\begin{equation}
	{\bm{{\cal A}}} = {\bm{{\cal U}}} * {\bm{{\cal S}}} * {{\bm{{\cal V}}}^T},
	\end{equation}
	where ${\bm{{\cal U}}} \in {{\mathbb{R}}^{{n_1} \times {n_1} \times {n_3}}}$, and ${\bm{{\cal V}}} \in {{\mathbb{R}}^{{n_2} \times {n_2} \times {n_3}}}$ are orthogonal tensors, ${\bm{{\cal S}}} \in {{\mathbb{R}}^{{n_1} \times {n_2} \times {n_3}}}$ is a f-diagonal tensor, frontal slices of which are all diagonal matrices. 
	
	The t-SVD based tensor nuclear norm ${{\left\|{\bm{{\cal A}}}\right\|}_ {\circledast} }$ employed in this paper can be defined as follows:
	\begin{equation}
	{\left\| {\bm{{\cal A}}} \right\|_ {\circledast} } = \sum\limits_{i = 1}^{\min ({n_1},{n_2})} {\sum\limits_{k = 1}^{{n_3}} {\left| {{{\bm{{\cal S}}}_f}_{iik}} \right|} },
	\end{equation}
	which has the following equivalence \cite{kilmer2013third}:
	\begin{equation}
	{\left\| {\bm{{\cal A}}} \right\|_ {\circledast} } = {\left\| {{\rm{bcirc}}({\bm{{\cal A}}})} \right\|_ * }.
	\end{equation}
	
	\subsection{Problem Formulation}
	Given multi-view data $\{ {\bm{X}^{(i)}}\} _{i = 1}^v$, under the self-expressiveness property that a data point can be effectively expressed by a linear combination of other data points in the same cluster or subspace, the subspace representation of the $i$-th view, i.e., ${\bm{Z}^{(i)}}$, can be learned from the following equation:
	\begin{equation}
	\label{subspace_representation}
	\mathop {\min }\limits_{{{\bm{Z}}^{(i)}}} {\kern 1pt} {\kern 1pt} {\kern 1pt} \lambda \Phi ({{\bm{X}}^{(i)}},{{\bm{X}}^{(i)}}{{\bm{Z}}^{(i)}}) + \Omega ({{\bm{Z}}^{(i)}}),
	\end{equation}
	in which $\Phi ( \cdot , \cdot )$ and $\Omega ( \cdot )$ denote the loss function and regularization term, respectively. For example, Sparse Subspace Clustering (SSC) \cite{elhamifar2013sparse} learns the subspace representation with the help of Frobenius norm and $l_1$ norm. Low-Rank Representation (LRR) leverages the nuclear norm and $l_{2,1}$ norm \cite{liu2012robust} for learning. Taking the LRR for example, The formulation of Eq.~(\ref{subspace_representation}) can be rewritten as follows:
	\begin{equation}
	\begin{array}{l}
	\mathop {\min }\limits_{{{\bm{Z}}^{(i)}},{{\bm{E}}^{(i)}}} {\kern 1pt} {\kern 1pt} {\kern 1pt} \lambda {\left\| {{{\bm{E}}^{(i)}}} \right\|_{2,1}} + {\left\| {{{\bm{Z}}^{(i)}}} \right\|_ * },{\kern 1pt} {\kern 1pt} {\kern 1pt} \\
	s.t.{\kern 1pt} {\kern 1pt} {\kern 1pt} {{\bm{X}}^{(i)}} = {{\bm{X}}^{(i)}}{{\bm{Z}}^{(i)}} + {{\bm{E}}^{(i)}},
	\end{array}
	\label{LRR_problem}
	\end{equation}
	and the following theorem can be achieved \cite{liu2012robust}:
	
	\newtheorem{thm}{\bf Theorem}[section]
	\begin{thm}
		\label{thm1}
		Assuming that samples of data are sufficient and the subspaces are independent, we can achieve that the rank of optimal ${\bm{Z}}^{(i)}$ of Eq.~(\ref{LRR_problem}) equals to the sum of dimensions of all subspaces.  
	\end{thm} 
	
	Most existing multi-view subspace clustering methods have the following framework:
	\begin{equation}
	\mathop {\min }\limits_{\left\{ {{{\bm{Z}}^{(i)}}} \right\}_{i = 1}^v} {\kern 1pt} {\kern 1pt} {\kern 1pt} \lambda \sum\limits_{i = 1}^v {\Phi ({{\bm{X}}^{(i)}},{{\bm{X}}^{(i)}}{{\bm{Z}}^{(i)}})}  + \sum\limits_{i = 1}^v {\Omega ({{\bm{Z}}^{(i)}})},
	\label{ExsitingModel}
	\end{equation}
	where some constraints are employing on the original subspace representations, i.e., ${\left\{ {{{\bm{Z}}^{(i)}}} \right\}_{i = 1}^v}$. 
	
\begin{figure}[!t]
		\centering
		\includegraphics[width=0.5\textwidth]{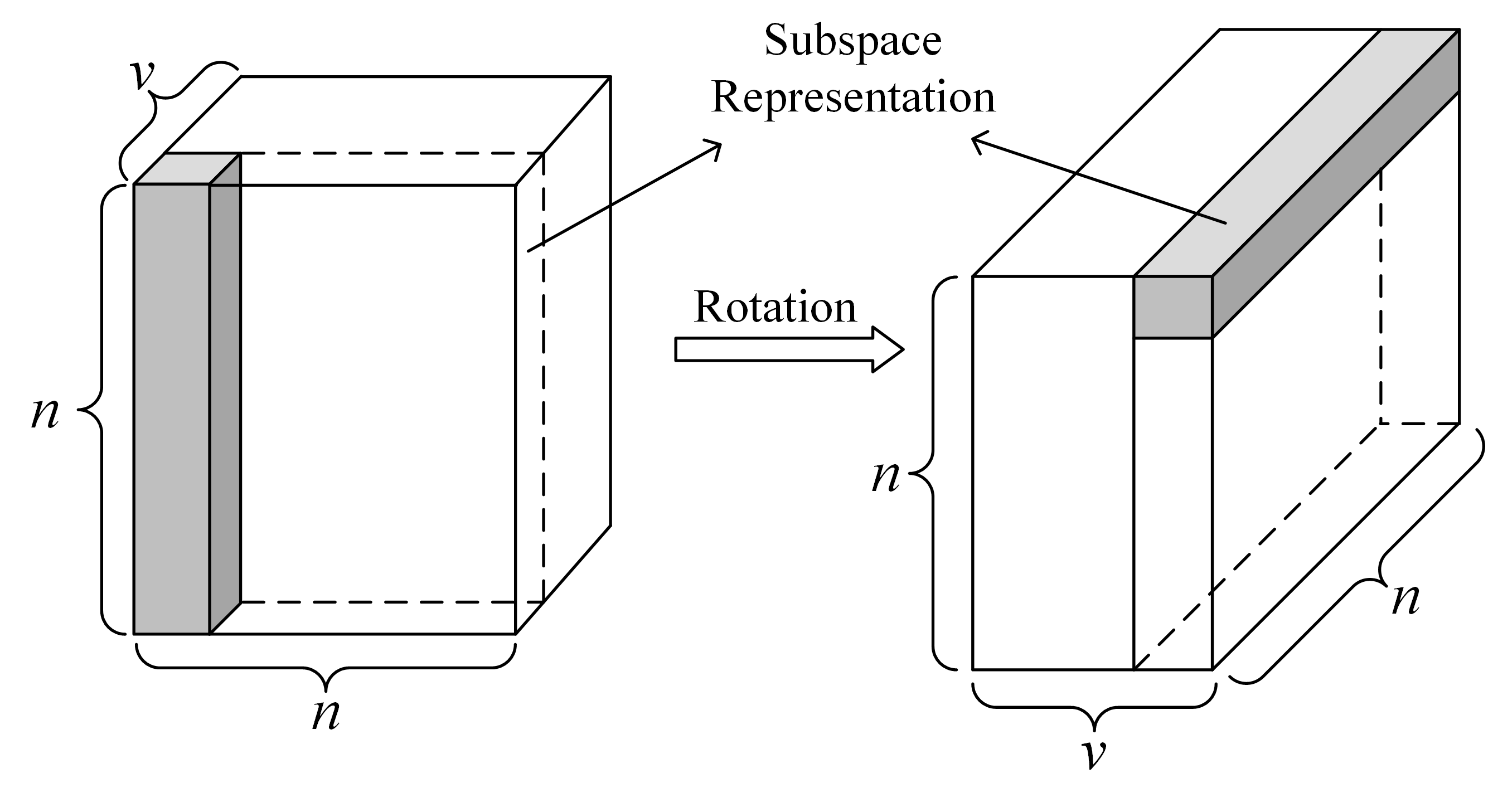}
		\caption{The way of tensor construction used in the proposed method. Based on the new subspace representations, i.e., $\{ {\bm{C}^{(i)}}\} _{i = 1}^v$, a tensor $\bm{{\cal C}}$ can be obtained and the rotation operation can be realized by the Matlab command $\rm{shiftdim}$ \cite{xie2018unifying}.}
		\label{TensorConstruction}
	\end{figure}

	As introduced in Section 2, for most existing multi-view subspace clustering approaches, they merely apply the specific constraint on these original subspace representation matrices directly. However, for multi-view data, the clustering properties of different views are usually diverse, even incompatible in real-life applications, since different views have their own specific statistic properties \cite{xu2013survey}. In other words, different views contain diverse statistic properties and could partially contradict with one another. Therefore, the way used in Eq.~(\ref{ExsitingModel}) ignores the specific information of multi-view data for the subspace representation learning. To address the aforementioned limitation, we introduce the rank-preserving decomposition, which are provided in Fig.~\ref{TISRL_Flowchart}. As can be derived from Theorem~\ref{thm1}, the rank of a subspace representation matrix is crucial to clustering. Here, we explain the reason behind the introduction of rank-preserving decomposition. To be specific, the rank-preserving decomposition conducted on the original subspace representation matrices can be formulated as follows:
	\begin{equation}
	{{\bm{Z}}^{(i)}} = {{\bm{P}}^{(i)}}{{\bm{C}}^{(i)}},{\kern 1pt} {\kern 1pt} {\kern 1pt} s.t.{\kern 1pt} {\kern 1pt} {\kern 1pt} {{\bm{P}}^{{{(i)}^T}}}{{\bm{P}}^{(i)}} = {\bm{I}},{\kern 1pt} {\kern 1pt} {\kern 1pt} i = 1,2, \cdots ,v,
	\label{LRD}
	\end{equation}
	where ${{\bm{P}}^{(i)}}$ denotes an orthogonal matrix, and ${{\bm{C}}^{(i)}}$ indicates a new subspace representation of the $i$-th view. Actually, based on Theorem~\ref{thm1}, it is easy to get that ${{\bm{C}}^{(i)}}$ and ${{\bm{Z}}^{(i)}}$ have the same nuclear norm and rank. According to Theorem~\ref{thm1}, the clustering properties of ${{\bm{Z}}^{(i)}}$ are preserved by ${{\bm{C}}^{(i)}}$, at the same time, the specific information of different views is embedded in ${{\bm{P}}^{(i)}}$ as well. Therefore, by introducing the rank-preserving decomposition, the aforementioned limitation of multi-view subspace clustering can be addressed elegantly. Subsequently, it is reasonable and natural to impose a specific regularization term on new subspace representation matrices for learning. In our method, we employ the t-SVD based low-rank tensor constraint, which can explore the high-order correlations of different views and is widely used in multi-view clustering methods \cite{xie2018unifying,wu2019essential}. To be specific, a 3-order tensor $\bm{{\cal C}}$ can be built as follows:
	\begin{equation}
	\bm{{\cal C}} = \varphi (\{{{\bm{C}}^{(1)}},{{\bm{C}}^{(2)}}, \cdots ,{{\bm{C}}^{(v)}}\}),
	\end{equation}
	where $\varphi(\cdot)$ builds the tensor $\bm{{\cal C}}$ by the way depicted in Fig.~\ref{TensorConstruction}. Subsequently, the t-SVD based low-rank tensor constraint can be applied on $\bm{{\cal C}}$.
	
	Therefore, as depicted in Fig.~\ref{TISRL_Flowchart}, the propose method takes Eq.~(\ref{LRD}) and Eq.~(\ref{TensorConstruction}) into consideration together, we model our TIRSL with the following objective function:
	\begin{equation}
	\begin{array}{l}
	\mathop {\min }\limits_{{{\bm{Z}}^{(i)}},{{\bm{C}}^{(i)}},{{\bm{P}}^{(i)}},{{\bm{E}}^{(i)}}} {\left\| {\bm{{\cal C}}} \right\|_ {\circledast} } + \lambda {\left\| {\bm{E}} \right\|_{2,1}},\\
	s.t.{\kern 1pt} {\kern 1pt} {\kern 1pt} {{\bm{X}}^{(i)}} = {{\bm{X}}^{(i)}}{{\bm{Z}}^{(i)}} + {{\bm{E}}^{(i)}},\\
	~~~~~~{{\bm{Z}}^{(i)}} = {{\bm{P}}^{(i)}}{{\bm{C}}^{(i)}},{\kern 1pt} {\kern 1pt} {\kern 1pt} {{\bm{P}}^{{{(i)}^T}}}{{\bm{P}}^{(i)}} = {\bm{I}},{\kern 1pt} {\kern 1pt} {\kern 1pt} i = 1,2, \cdots ,v,\\
	~~~~~~{\bm{{\cal C}}} = \varphi (\{ {{\bm{C}}^{(1)}},{{\bm{C}}^{(2)}}, \cdots ,{{\bm{C}}^{(v)}}\} ),\\
	~~~~~~{\bm{E}} = [{{\bm{E}}^{(1)}};{{\bm{E}}^{(2)}}; \cdots ;{{\bm{E}}^{(v)}}],
	\end{array}
	\label{Obj_Func}
	\end{equation}
	where $\lambda$ indicates a tradeoff parameter. For the matrix $\bm{E}$, as recommended  in \cite{gLMSC2020PAMI,lang2012saliency}, all the error matrices of multiple views, i.e., $\{ {\bm{E}^{(i)}}\} _{i = 1}^v$, are jointly concatenates vertically along the column direction, and the $l_{2,1}$-norm is employed in this objective function to force columns of $\bm{E}$ to be more close to zeros \cite{liu2012robust}. It can be also observed that no extra hyper-parameter is added by introducing the rank preserving decomposition in the proposed TISRL.
	
	When the proposed objective function is optimized, the desired intrinsic subspace representation of multi-view data, i.e., ${\bm{C}}$, can be calculated as follows:
	\begin{equation}
	{\bm{C}} = \frac{1}{v}\sum\limits_{i = 1}^v {({{\left| {\bm{C}} \right|}^{(i)}} + {{\left| {\bm{C}} \right|}^{{{(i)}^T}}})}.
	\end{equation}
	To get clustering results, the standard spectral clustering algorithm \cite{von2007tutorial} can be performed on the learned ${\bm{C}}$.
	
	\subsection{Optimization}
    To solve the proposed objective function, i.e., Eq.~(\ref{Obj_Func}), an alternating direction minimization algorithm based on the ALM \cite{lin2011linearized} is designed in this section. We briefly summary the whole optimization process in Algorithm 1. Concretely, the corresponding augmented Lagrangian function of Eq.~(\ref{Obj_Func}) can be modeled as follows:
	\begin{equation}
	\begin{array}{l}
	{{\cal L}}(\{ {{\bm{C}}^{(i)}},{{\bm{P}}^{(i)}},{{\bm{Z}}^{(i)}},{{\bm{E}}^{(i)}},{\bm{Y_x}}^{(i)},{\bm{Y_z}}^{(i)}\} _{i = 1}^v,{\kern 1pt} \bm{{\cal Q}},\bm{{\cal W}})\\
	{\kern 1pt} {\kern 1pt} {\kern 1pt} {\kern 1pt} {\kern 1pt} {\kern 1pt} {\kern 1pt} {\kern 1pt} {\kern 1pt}  = {\left\| \bm{{\cal Q}} \right\|_ {\circledast} } + \lambda {\left\| {\bm{E}} \right\|_{2,1}} + {\Psi _1}(\bm{{\cal W}},\bm{{\cal C}} - \bm{{\cal Q}})\\
	~~~~~~~~~  + \sum\limits_{i = 1}^v {{\Psi _2}({\bm{Y_x}}^{(i)},{{\bm{X}}^{(i)}} - {{\bm{X}}^{(i)}}{{\bm{Z}}^{(i)}} - {{\bm{E}}^{(i)}})} \\
	~~~~~~~~~  + \sum\limits_{i = 1}^v {{\Psi _2}({\bm{Y_z}}^{(i)},{{\bm{Z}}^{(i)}} - {{\bm{P}}^{(i)}}{{\bm{C}}^{(i)}})}, \\
	s.t. ~~~ {{\bm{P}}^{{{(i)}^T}}}{{\bm{P}}^{(i)}} = {\bm{I}},
	\end{array}
	\end{equation}
	where $\bm{{\cal Q}}$ is an introduced auxiliary tensor variable, $\{{\bm{Y_x^{(i)}}}\}_{i = 1}^v$, $\{{\bm{Y_z^{(i)}}}\}_{i = 1}^v$, and $\bm{{\cal W}}$ are all Lagrange multipliers. For ${\Psi _1}( \cdot,\cdot )$ and ${\Psi _2}( \cdot,\cdot )$, they have the following formulas:
	\begin{equation}
	\begin{array}{l}
	{\Psi _1}({\bm{{\cal A}}},{\bm{{\cal B}}}) = \left\langle {{\bm{{\cal A}}},{\bm{{\cal B}}}} \right\rangle  + \frac{\rho }{2}\left\| {\bm{{\cal B}}} \right\|_F^2,\\
	{\Psi _2}({\bm{A}},{\bm{B}}) = \left\langle {{\bm{A}},{\bm{B}}} \right\rangle  + \frac{\mu }{2}\left\| {\bm{B}} \right\|_F^2,
	\end{array}
	\end{equation}
	where $\rho$ and $\mu$ denote the penalty parameters of optimization process \cite{lin2011linearized}.
	
	\begin{algorithm}[t]
		\caption{Algorithm of the proposed TISRL}
		\hspace*{0.06in} {\bf Input:} \\
		\hspace*{0.18in} Multi-view $\{ {{\bm{X}}^{(i)}}\} _{i = 1}^v$, tradeoff parameter $\lambda$\\
		\hspace*{0.06in} {\bf Output:} \\ 
		\hspace*{0.18in} Intrinsic subspace representation: ${\bm{C}}$,\\
		\hspace*{0.06in} {\bf Initialization:} \\
		\hspace*{0.18in} {\bf For} $i = 1,2 \cdots ,v$ {\bf do}:\\
		\hspace*{0.36in} ${\bm{Z}}^{(i)}={\bm{0}}$, ${\bm{E}}^{(i)}={\bm{0}}$, ${\bm{P}}^{(i)}={\bm{0}}$,
		${\bm{C}}^{(i)}={\bm{0}}$, ${\bm{Y_x}}^{(i)}={\bm{0}}$, ${\bm{Y_z}}^{(i)}={\bm{0}}$,\\
		\hspace*{0.18in} {\bf End}\\
		\hspace*{0.18in} ${\bm{{\cal Q}}}={\bm{0}}$, ${\bm{{\cal W}}}={\bm{0}}$, $\varepsilon={10^{ - 7}}$,
		$\mu=\rho={10^{ - 5}}$, $\eta=2.0$, $\mu_{max}=\rho_{max}={10^{ 12}}$.\\
		\hspace*{0.06in} {\bf Optimization:}\\
		\hspace*{0.18in} {\bf Repeat:}\\
		\hspace*{0.36in} {\bf For} $i = 1,2 \cdots ,v$ {\bf do}:\\
		\hspace*{0.60in} Updating ${\bm{Z}}^{(i)}$ by leveraging Eq.~(19);\\
		\hspace*{0.60in} Updating ${\bm{P}}^{(i)}$ by soloving Eq.~(21);\\
		\hspace*{0.60in} Updating ${\bm{E}}^{(i)}$ by soloving Eq.~(24);\\
		\hspace*{0.60in} Updating ${\bm{C}}^{(i)}$ by leveraging Eq.~(26);\\
		\hspace*{0.60in} Updating ${\bm{Y}}_x^{(i)}$ by leveraging Eq.~(32);\\
		\hspace*{0.60in} Updating ${\bm{Y}}_z^{(i)}$ by leveraging Eq.~(32);\\
		\hspace*{0.36in} {\bf End}\\
		\hspace*{0.36in} Updating ${\bm{{\cal Q}}}$ by leveraging Eq.~(28);\\
		\hspace*{0.36in} Updating ${\bm{{\cal W}}}$ by leveraging Eq.~(32);\\
		\hspace*{0.36in} Updating $\mu$ and $\rho$ by leveraging Eq.~(32);\\
		\hspace*{0.18in} {\bf Until:}\\
		\hspace*{0.36in} {\bf For} $i = 1,2 \cdots ,v$ {\bf do}:\\
		\hspace*{0.60in} ${\left\| {{{\bm{X}}^{(i)}} - {{\bm{X}}^{(i)}}{{\bm{Z}}^{(i)}} - {\bm{E}}^{(i)}} \right\|_\infty } < \varepsilon, $\\
		\hspace*{0.60in} ${\left\| {{{\bm{Z}}^{(i)}} - {P^{(i)}}{\bm{C}}^{(i)}} \right\|_\infty } < \varepsilon, $\\
		\hspace*{0.60in} ${\left\| {{{\bm{Q}}^{(i)}} - {W^{(i)}}} \right\|_\infty } < \varepsilon, $ and \\
		\hspace*{0.36in} {\bf End}\\
		\hspace*{0.18in} ${\bm{C}} = \frac{1}{v}\sum\limits_{i = 1}^v {({{\left| {\bm{C}} \right|}^{(i)}} + {{\left| {\bm{C}} \right|}^{{{(i)}^T}}})}$.
	\end{algorithm}

	1) Updating $\{{\bm{Z}}^{(i)}\}_{i = 1}^v$. 
	To update $\{{\bm{Z}}^{(i)}\}_{i = 1}^v$, other variables should be fixed in the augmented Lagrangian function, and then the corresponding subproblem of updating ${\bm{Z}}^{(i)}$ can be obtained. We formulate is as follows:
	\begin{equation}
	\begin{array}{l}
	\mathop {\min }\limits_{{{\bm{Z}}^{(i)}}} {\kern 1pt} {\kern 1pt} {\kern 1pt} {\Psi _2}({\bm{Y_x}}^{(i)},{{\bm{X}}^{(i)}} - {{\bm{X}}^{(i)}}{{\bm{Z}}^{(i)}} - {{\bm{E}}^{(i)}}) + {\Psi _2}({\bm{Y_z}}^{(i)},{{\bm{Z}}^{(i)}} - {{\bm{\textsc{P}}}^{(i)}}{{\bm{C}}^{(i)}}).
	\end{array}
	\label{Z_subproblem}
	\end{equation}
	Differentiating the function in (\ref{Z_subproblem}) w.r.t. ${\bm{Z}}^{(i)}$ and then letting it to zero, the optimal ${\bm{Z}}^{(i)}$ can be attained as follows:
	\begin{equation}
	\begin{array}{l}
	{{\bm{Z}}^{(i)}} = {(\mu {\bm{I}} + \mu {{\bm{X}}^{{{(i)}^T}}}{{\bm{X}}^{(i)}})^{ - 1}}({{\bm{X}}^{{{(i)}^T}}}{\bm{Y_x}}^{(i)}\\
	+ \mu {{\bm{X}}^{{{(i)}^T}}}{{\bm{X}}^{(i)}} - \mu {{\bm{X}}^{{{(i)}^T}}}{{\bm{E}}^{(i)}} - {\bm{Y_z}}^{(i)} + \mu {{\bm{P}}^{(i)}}{{\bm{C}}^{(i)}})
	\end{array}
	\label{Z_optimization}
	\end{equation}
	
	2) Updating $\{{\bm{P}}^{(i)}\}_{i = 1}^v$.
	Similarly, the subproblem of updating ${\bm{P}}^{(i)}$ can be also obtained by fixing other variables. We can write it as follows:
	\begin{equation}
	\begin{array}{l}
	\mathop {\min }\limits_{{{\bm{P}}^{(i)}}} {\kern 1pt} {\kern 1pt} {\kern 1pt} \left\langle {{\bm{Y_z}}^{(i)},{{\bm{Z}}^{(i)}} - {{\bm{P}}^{(i)}}{{\bm{C}}^{(i)}}} \right\rangle  + \frac{\mu }{2}\left\| {{{\bm{Z}}^{(i)}} - {{\bm{P}}^{(i)}}{{\bm{C}}^{(i)}}} \right\|_F^2\\
	s.t.{\kern 1pt} {\kern 1pt} {\kern 1pt} {{\bm{P}}^{{{(i)}^T}}}{{\bm{P}}^{(i)}} = {\bm{I}},
	\end{array}
	\label{P_subproblem_orignal}
	\end{equation}
	which can be remodeled as follows:
	\begin{equation}
	\begin{array}{l}
	\mathop {\min }\limits_{{{\bm{P}}^{(i)}}} {\kern 1pt} {\kern 1pt} {\kern 1pt} \frac{\mu }{2}\left\| {({{\bm{Z}}^{(i)}} + \frac{1}{\mu }{\bm{Y_z}}^{(i)}) - {{\bm{P}}^{(i)}}{{\bm{C}}^{(i)}}} \right\|_F^2,\\
	s.t.{\kern 1pt} {\kern 1pt} {\kern 1pt} {{\bm{P}}^{{{(i)}^T}}}{{\bm{P}}^{(i)}} = {\bm{I}}.
	\end{array}
	\label{P_subproblem}
	\end{equation}
	which is an orthogonal procrustes problem. For optimization, we have the following theorem:
	\begin{thm}
		\label{thm3}
		\cite{gower2004procrustes} The goal of Eq.~(\ref{P_subproblem}) is to obtain the nearest orthogonal matrix to a given matrix, i.e., $({{\bm{Z}}^{(i)}} + \frac{1}{\mu }{\bm{Y_z}}^{(i)}){{\bm{C}}^{{{(i)}^T}}}$, and the optimal solution can be written as follows: 
		\begin{equation}
		{{\bm{P}}^{(i)}} = {{\bm{U}}^{(i)}}{{\bm{V}}^{{{(i)}^T}}},
		\end{equation}
		in which ${{\bm{U}}^{(i)}}$ and ${{\bm{V}}^{{{(i)}^T}}}$ can be calculated by leveraging the following SVD:
		\begin{equation}
		({{\bm{Z}}^{(i)}} + \frac{1}{\mu }{\bm{Y_z}}^{(i)}){{\bm{C}}^{{{(i)}^T}}} = {{\bm{U}}^{(i)}}{{\bm{\Sigma}} ^{(i)}}{{\bm{V}}^{{{(i)}^T}}}.
		\end{equation}
	\end{thm} 
	Therefore, based on the {\bf{Theorem~\ref{thm3}}}, the optimal result of Eq.~(\ref{P_subproblem_orignal}) can be obtained directly.
	
	3) Updating $\{{\bm{E}}^{(i)}\}_{i = 1}^v$. For the updating of $\{{\bm{E}}^{(i)}\}_{i = 1}^v$, we written the corresponding subproblem as follows when other variables are fixed:
	\begin{equation}
	\lambda {\left\| {\bm{E}} \right\|_{2,1}} + \frac{\mu }{2}\left\| {{\bm{E}} - \Gamma (\{ {{\bm{X}}^{(i)}} - {{\bm{X}}^{(i)}}{{\bm{Z}}^{(i)}} - \frac{1}{\mu }{\bm{Y_x}}^{(i)}\} _{i = 1}^v)} \right\|_F^2,
	\label{E_subproblem}
	\end{equation}
	where $\Gamma (\cdot)$ is a function that concatenates $\{ {{\bm{X}}^{(i)}} - {{\bm{X}}^{(i)}}{{\bm{Z}}^{(i)}} - \frac{1}{\mu }{\bm{Y_x}}^{(i)}\} _{i = 1}^v$ vertically. By using Lemma 3.2 in \cite{liu2012robust}, the closed solution of Eq.~(\ref{E_subproblem}) can be obtained effectively. Once $E$ is optimized, $\{{\bm{E}}^{(i)}\}_{i = 1}^v$ can obtained straightforward.
	
	4) Updating $\{{\bm{C}}^{(i)}\}_{i = 1}^v$.
	To update ${\bm{C}}^{(i)}$, we fix other variables and formulate the following problem:
	\begin{equation}
	\begin{array}{l}
	\mathop {\min }\limits_ { {\bm{C}}^{(i)}} {\kern 1pt} {\kern 1pt} {\kern 1pt} \left\langle {{{\bm{W}}^{(i)}},{ {\bm{C}}^{(i)}} - {{\bm{Q}}^{(i)}}} \right\rangle  + \frac{\rho }{2}\left\| {{{\bm{C}}^{(i)}} - {{\bm{Q}}^{(i)}}} \right\|_F^2\\
	~~~~~~~~+ \left\langle {{\bm{Y_z}}^{(i)},{{\bm{Z}}^{(i)}} - {{\bm{P}}^{(i)}}{{\bm{C}}^{(i)}}} \right\rangle  + \frac{\mu }{2}\left\| {{{\bm{Z}}^{(i)}} - {{\bm{P}}^{(i)}}{{\bm{C}}^{(i)}}} \right\|_F^2
	\end{array}
	\label{C_subproblem}
	\end{equation}
	where ${ {\bm{C}}^{(i)}}$ and ${ {\bm{W}}^{(i)}}$ can be attained by separately performing the inverse operation of $\varphi ( \cdot )$ on $\bm{{\cal C}}$ and $\bm{{\cal W}}$, then extracting the $i$-th frontal slice.
	
	Similar to the solution in the subproblem of updating ${\bm{Z}}^{(i)}$, we can derivate the function in Eq.~(\ref{C_subproblem}) and then let it to zero. We get the following optimal result here:
	\begin{equation}
	\begin{array}{l}
	{{\bm{C}}^{(i)}} = {(\rho {\bm{I}} + \mu {{\bm{P}}^{{{(i)}^T}}}{{\bm{P}}^{(i)}})^{ - 1}}(\rho {{\bm{Q}}^{(i)}}
	- {{\bm{W}}^{(i)}} + {{\bm{P}}^{{{(i)}^T}}}{\bm{Y_z}}^{(i)} + \mu {{\bm{P}}^{{{(i)}^T}}}{{\bm{Z}}^{(i)}}).
	\end{array}
	\label{C_optimization}
	\end{equation}
	
	5) Updating $\bm{{\cal Q}}$.
	With other variables being fixed, we get the following subproblem of updating $\bm{{\cal Q}}$:
	\begin{equation}
	\mathop {\min }\limits_{\bm{{\cal Q}}} {\kern 1pt} {\kern 1pt} {\kern 1pt} {\left\| {\bm{{\cal Q}}} \right\|_ {\circledast} } + \frac{\rho }{2}\left\| {{\bm{{\cal Q}}} - ({\bm{{\cal C}}} + \frac{1}{\rho }{\bm{{\cal W}}})} \right\|_F^2
	\end{equation}
	which can be optimized as follows \cite{hu2016twist}:
	\begin{equation}
	{\bm{{\cal Q}}} = {\bm{{\cal U}}} * {{\bm{{\cal F}}}_{\rho {\rm{n}}}}({\bm{{\cal S}}}) * {{\bm{{\cal V}}}^T},
	\label{Q_Optimization}
	\end{equation}
	in which ${\bm{{\cal U}}}$, ${\bm{{\cal S}}}$, and ${\bm{{\cal V}}}$ can be calculated as follows:
	\begin{equation}
	{\bm{{\cal U}}} * {\bm{{\cal S}}} * {{\bm{{\cal V}}}^T} = {\bm{{\cal C}}} + \frac{1}{\rho }{\bm{{\cal W}}},
	\end{equation}
	and ${{\bm{{\cal F}}}_{\rho {\rm{n}}}}({\bm{{\cal S}}})$ has the following formulation:
	\begin{equation}
	{{\bm{{\cal F}}}_{\rho {\rm{n}}}}({\bm{{\cal S}}})={\bm{{\cal S}}} * {\bm{{\cal J}}},
	\end{equation}
	where ${\bm{{\cal J}}} \in {{\mathbb{R}}^{n \times v \times n}}$ is f-diagonal tensor with the following elements after performing ${\rm{fft}({\bm{{\cal J}}},[~],3)}$:
	\begin{equation}
	{{\bm{{\cal J}}}_f}(i,i,k) = \max (1 - \frac{{\rho n}}{{{\bm{{\cal S}}}_f^{(k)}(i,i)}},0).
	\end{equation}
	
	6) Updating $\mu$, $\rho$, $\{{\bm{Y_x}}^{(i)}\}_{i = 1}^v$, $\{{\bm{Y_z}}^{(i)}\}_{i = 1}^v$, and $\bm{{\cal Q}}$.
	According \cite{lin2011linearized}, penalty parameters and Lagrange multipliers can be updated as follows:
	\begin{equation}
	\begin{array}{l}
	\mu  = \min (\eta \mu ,{\mu _{\max }}),\\
	\rho  = \min (\eta \rho ,{\rho _{\max }}),\\
	{\bm{Y_x}}^{(i)} = {\bm{Y_x}}^{(i)} + \mu ({{\bm{X}}^{(i)}} - {{\bm{X}}^{(i)}}{{\bm{Z}}^{(i)}} - {{\bm{E}}^{(i)}}),\\
	{\bm{Y_z}}^{(i)} = {\bm{Y_z}}^{(i)} + \mu ({{\bm{Z}}^{(i)}} - {{\bm{P}}^{(i)}}{{\bm{C}}^{(i)}}),\\
	{\bm{{\cal W}}} = {\bm{{\cal W}}} + \rho ({\bm{{\cal C}}} - {\bm{{\cal Q}}})
	\end{array}
	\label{LagrangeMultipler_Optimization}
	\end{equation}
	where $\eta$ is a scale that is utilized to monotonically increase $\mu$ and $\rho$ until reaching ${\mu _{\max }}$ and ${\rho _{\max }}$, respectively.

	\subsection{Convergence Analysis}
	As can be observed in Algorithm 1, more than two subproblems are involved in our algorithm. Therefore, it is difficult to prove its convergence \cite{lin2011linearized}. Fortunately, experimental results in the following section demonstrate that the proposed algorithm achieves convergence quickly and stably for all given real-life multi-view datasets used in this paper.
	
	\section{Experiments}
	In this section, we carry on experiments on nine real-life multi-view datasets, and We also report the corresponding experimental results and analyses in this section. To be clear, experiments are run on a personal computer with 8 cores 2.10GHz Intel Xeon CPU and 128GB RAM.
	
	\begin{table}[!t]
		\caption{Statistic information of nine multi-view datasets.}
		\label{Datasets_Statistic}
		\centering
		\resizebox{0.45\textwidth}{!}{
			\begin{tabular}{l|c|c|c|c}
				\toprule
				Dataset & Objective & \# Samples & \# Clusters & \# Views\\
				\midrule
				BBCSport & Text & 544 & 5 & 2  \\
				\midrule
				NGs & Text & 500 & 5 & 3  \\
				\midrule
				Caltech-101 & Object & 8677 & 101 & 4  \\
				\midrule
				Caltech-7 & Object & 441 & 7 & 3  \\
				\midrule
				MSRCV1 & Object  & 210 & 7 & 6  \\
				\midrule
				Notting-Hill & Face & 4660 & 5 & 3  \\
				\midrule
				ORL & Face & 400 & 40 & 3  \\
				\midrule
				MITIndoor-67 & Scene & 5360 & 67 & 4  \\
				\midrule
				Scene-15 & Scene & 4485 & 15 & 3  \\
				\bottomrule
		\end{tabular}}
	\end{table}
	
	\begin{figure*}[!t]
		\centering
		\includegraphics[width=1\textwidth]{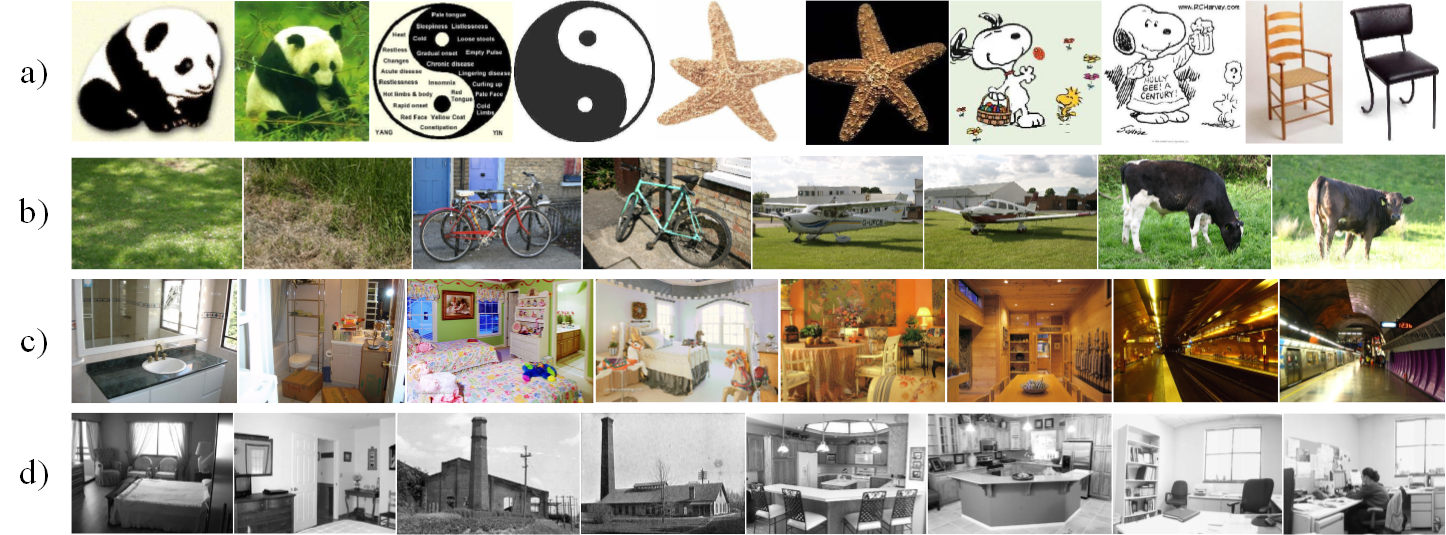}
		\caption{Illustration of some example images selected from a) Caltech-101, b) MSRCV1, c) MITIndoor-67, and d) Scene-15. For Caltech-101 and MSRCV1, they are objective image dataset. Regarding to MITIndoor-67 and Scene-15, they are scene image dataset.}
		\label{Data_Example}
	\end{figure*}
	
	\subsection{Experimental Settings}
	Nine real-world multi-view datasets are employed in this section. Some statistics of these real-life benchmark datasets are summarized in Table~\ref{Datasets_Statistic} and some samples of images datasets are presented in Fig.~\ref{Data_Example}. They are introduced as follows:
	
	\begin{enumerate}
		\item BBCsport\footnote{http://mlg.ucd.ie/datasets/} \cite{greene2006practical}, which is a news multi-view dataset, is composed of 544 sample points and 2 views.
		\item NGs\footnote{http://lig-membres.imag.fr/grimal/data.html} \cite{hussain2010icmla} is also a new multi-view datasets and has 500 sample points and 3 types of features.  
		\item Caltech-101\footnote{http://www.vision.caltech.edu/Image\_Datasets/Caltech101/} \cite{fei2004learning} is a challenging objective image multi-view datasets. It contains 8677  sample points from 101 different clusters, and 4 different views, including three traditional features (PHOW \cite{bosch2007image}, LBP \cite{ojala2002multiresolution}, CENTRIST \cite{wu2010centrist}) and a deep features \cite{szegedy2016rethinking}.
		\item Caltech-7 is a challenging objective image multi-view datasets with 441 sample and 3 views. Three types (PHOW, LBP, and CENTRIST) are involved in these dataset as three views.
		\item MSRCV1\footnote{http://research.microsoft.com/en-us/projects/objectclassrecognition/} \cite{xu2016discriminatively} contains 210 samples and 6 different feature typesis, and the image samples of this dataset are collects 7 objective categories.
		\item Notting-Hill\cite{zhang2009character} is a face dataset and 3 features, including Intensity, LBP and Gabor, are used in experiments, and it contains 4660 samples from 5 individuals.
		\item ORL\footnote{http://www.uk.research.att.com/facedatabase.html} is composed of 400 face images from 40 different individual, and the same to Notting-Hill, Intensity, LBP and Gabor are employed as 3 different views.
		\item MITIndoor-67 \cite{quattoni2009recognizing} is a scene multi-view dataset, which has 5360 data samples from 67 categories, and 4 features, including PHOW, LBP, CENTRIST, and deep features extracted from a pre-training VGG-VD network \cite{simonyan2014very} are utilized in experiments.
		\item Scene-15 \cite{fei2005bayesian} is composed of 4485 images collected from 15 different natural scene categories, and PHOW, LBP, and CENTRIST are leveraged as 3 different views. 	
	\end{enumerate}
	
	In this section, 10 state-of-the-art clustering approaches, including 2 single view clustering methods and 8 multi-view clustering methods are leveraged in this section for comparison. Specifically, 3 low-rank tensor based multi-view clustering algorithms, including LT-MSC \cite{zhang2015low}, t-SVD-MSC \cite{xie2018unifying}, and ETLMSC \cite{wu2019essential}, are employed here. We briefly introduce these methods as follows:
	
	\begin{enumerate}
		\item SPC$\rm{_{best}}$ \cite{von2007tutorial} is the standard spectral clustering, which is applied on each views and the best clustering results are presented here.
		\item LRR$\rm{_{best}}$ \cite{liu2012robust} is a clustering algorithm based on low-rank representation. It is employed on each views and the best clustering results are presented here. 
		\item RMSC \cite{xia2014robust} is a robust multi-view spectral clustering, it achieves the multi-view clustering by pursuing a low-rank transition probability matrices of all views.
		\item AMGL \cite{nie2016parameter} is the auto-weighted multiple graph learning framework, which can be used for multi-view clustering.
		\item GMC \cite{wang2019gmc} is a graph-based multi-view clustering, which takes graph matrices of multiple views into consideration and learns a fused graph for clustering.
		\item DiMSC \cite{cao2015diversity} is a diversity-induced multi-view subspace clustering. The Hilbert Schmidt Independence Criterion is leveraged to explore the complementarity of multi-view representations.
		\item LMSC \cite{gLMSC2020PAMI} is a latent multi-view subspace clustering, it aims to learn a latent representation and a corresponding subspace representation simultaneously for clustering.
		\item LT-MSC \cite{zhang2015low} is a low-rank tensor constrained multiview subspace clustering, which leverages a generalized tensor nuclear norm on the original subspace representations. 
		\item t-SVD-MSC \cite{xie2018unifying} is a t-SVD-based multi-view subspace clustering method, which employs a t-SVD-based tensor nuclear norm on the original subspace representations.
		\item ETLMSC \cite{wu2019essential} is an essential tensor learning for multi-view spectral clustering, which imposes a t-SVD-based nuclear norm on the transition probability matrices of all views to learn the essential tensor for clustering.
	\end{enumerate}
	
	Furthermore, 4 evaluation metrics are also employed in this section. Specifically, NMI, ACC, F-score, and Precision \cite{zhang2015low} are leveraged to present the clustering performance quantitatively. Each algorithm is run 30 times and average results are reported in this section, and the best and second best results are highlighted in bold font and underline respectively. Moreover, it should be noted that only the average results are reported and the standard deviations do not be provided in this section, since the standard deviations under all metrics are smaller than 0.1 on all datasets.
	
	\begin{table*}[!t]
		\caption{Experimental results of different methods on BBCSport, NGs, and Caltech-101. For the proposed TISRL, lambda are fixed to $0.10$, $0.01$ and $0.01$, respectively.}
		\label{Results_all_1}
		\centering
		\resizebox{0.95\textwidth}{!}{
			\begin{tabular}{l|c|c|c|c||c|c|c|c||c|c|c|c}
				\toprule
				Dataset & \multicolumn{4}{c||}{BBCSport} & \multicolumn{4}{c||}{NGs} & \multicolumn{4}{c}{Caltech-101}\\
				\midrule
				Method & NMI & ACC & F-score & Precision & NMI & ACC & F-score & Precision & NMI & ACC & F-score & Precision \\
				\midrule
				SPC$_{\rm{best}}$ & 0.735 & 0.853 & 0.798 & 0.804 & 0.016 & 0.204 & 0.330 & 0.198 & 0.723 & 0.484 & 0.340 & 0.597 \\
				\midrule
				LRR$_{\rm{best}}$ & 0.747 & 0.886  & 0.789  & 0.803  & 0.340  & 0.421  & 0.391 & 0.269 & 0.728 & 0.510 & 0.339 & 0.627 \\
				\midrule
				RMSC & 0.808 & 0.912 & 0.871 & 0.879 & 0.158 & 0.370 & 0.370 & 0.266 & 0.573 & 0.346 & 0.258 & 0.457 \\
				\midrule
				AMGL & 0.864 & 0.919 & 0.901 & 0.871 & 0.898 & 0.939 & 0.921 & 0.909 & 0.440 & 0.232 & 0.061 & 0.034 \\
				\midrule
				GMC & 0.795 & 0.739 & 0.721 & 0.573 & 0.939 & 0.982 & 0.964 & 0.964 & 0.544 & 0.331 & 0.081 & 0.044 \\
				\midrule
				DiMSC & 0.814 & 0.901 & 0.880 & 0.875 & 0.819 & 0.826 & 0.797 & 0.759 & 0.589 & 0.351 & 0.253 & 0.362 \\
				\midrule
				LMSC & 0.783 & 0.858 & 0.762 & 0.757 & 0.905 & 0.971 & 0.942 & 0.942 & 0.818 & 0.572 & 0.387 & 0.694 \\
				\midrule
				LT-MSC & 0.066 & 0.379 & 0.383 & 0.239  & 0.965 & 0.990 & 0.980 & 0.980 & 0.788 & 0.559 & 0.403 & 0.670 \\
				\midrule
				t-SVD-MSC & 0.830 & 0.941  & 0.888 & 0.881 & \underline{0.972} & \underline{0.992} & \underline{0.984} & \underline{0.984} & 0.858 & 0.607 & 0.440 & 0.742 \\
				\midrule
				ETLMSC & \underline{0.984} & \underline{0.978}  & \underline{0.977}  & \underline{0.963}  & 0.601 & 0.656 & 0.623 & 0.578 & \underline{0.899} & \underline{0.639} & \underline{0.465} & \bf{0.825} \\
				\midrule
				TISRL & \bf{1.000} & \bf{1.000}  & \bf{1.000}  & \bf{1.000}  & \bf{1.000} & \bf{1.000} & \bf{1.000} & \bf{1.000} & \bf{0.900} & \bf{0.667} & \bf{0.490} & \underline{0.816} \\
				\bottomrule
		\end{tabular}}
	\end{table*}
	
	\begin{table*}[!t]
		\caption{Experimental results of different methods on Caltech-7, MSRCV1, and Notting-Hill. For the proposed TISRL, lambda are fixed to $0.002$, $0.70$ and $0.01$, respectively.}
		\label{Results_all_2}
		\centering
		\resizebox{0.95\textwidth}{!}{
			\begin{tabular}{l|c|c|c|c||c|c|c|c||c|c|c|c}
				\toprule
				Dataset & \multicolumn{4}{c||}{Caltech-7} & \multicolumn{4}{c||}{MSRCV1} & \multicolumn{4}{c}{Notting-Hill}\\
				\midrule
				Method & NMI & ACC & F-score & Precision & NMI & ACC & F-score & Precision & NMI & ACC & F-score & Precision \\
				\midrule
				SPC$_{\rm{best}}$ & 0.440 & 0.528 & 0.474 & 0.451 & 0.605 & 0.683 & 0.572 &0.563 & 0.723 & 0.816 & 0.775 & 0.780 \\
				\midrule
				LRR$_{\rm{best}}$ & 0.443 & 0.549 & 0.500 & 0.469 & 0.570 & 0.674 & 0.536 & 0.529 & 0.579 & 0.794 & 0.653 & 0.672\\
				\midrule
				RMSC & 0.374 & 0.510 & 0.417 & 0.436 & 0.673 & 0.789 & 0.666 & 0.656 & 0.585 & 0.807 & 0.603 & 0.621 \\
				\midrule
				AMGL & 0.393 & 0.472 & 0.365 & 0.264 & 0.736 & 0.717 & 0.645 &0.569 & 0.129 & 0.358 & 0.369 & 0.230\\
				\midrule
				GMC & 0.424 & 0.479 & 0.356 & 0.243 & 0.820 & 0.895 & 0.800 & 0.786 & 0.092 & 0.312 & 0.369 & 0.228\\
				\midrule
				DiMSC & 0.336 & 0.493 & 0.375 & 0.389 & 0.632 & 0.719 & 0.612 & 0.596 & 0.799 & 0.837 & 0.834 & 0.822 \\
				\midrule
				LMSC & 0.437 & 0.518 & 0.452 & 0.434 & 0.615 & 0.695 & 0.591 & 0.573 & 0.697 & 0.816 & 0.761 & 0.782\\
				\midrule
				LT-MSC &  0.455 & 0.533 & 0.486 & 0.466 & 0.756 & 0.843 & 0.737 & 0.726 & 0.779 & 0.868 & 0.825 & 0.830\\
				\midrule
				t-SVD-MSC & \underline{0.801} & \underline{0.775} & \underline{0.826} & \underline{0.852} & 0.729 & 0.858 & 0.782 & 0.766 & 0.900 & \underline{0.957} & 0.922 & 0.937 \\
				\midrule
				ETLMSC & 0.631 & 0.680 & 0.655 & 0.653 & \underline{0.878} & \underline{0.888} & \underline{0.849} & \underline{0.834} & \underline{0.911} & 0.951 & \underline{0.924} & \underline{0.940} \\
				\midrule
				TISRL & \bf{0.834} & \bf{0.812} & \bf{0.829} & \bf{0.858} & \bf{1.000} & \bf{1.000} & \bf{1.000} & \bf{1.000} & \bf{0.940} & \bf{0.979} & \bf{0.962} & \bf{0.970} \\
				\bottomrule
		\end{tabular}}
	\end{table*}
	
	\begin{table*}[!t]
		\caption{Experimental results of different methods on ORL, MITIndoor-67, and Scene-15. For the proposed TISRL, lambda are fixed to $0.20$, $0.05$ and $0.005$, respectively.}
		\label{Results_all_3}
		\centering
		\resizebox{0.95\textwidth}{!}{
			\begin{tabular}{l|c|c|c|c||c|c|c|c||c|c|c|c}
				\toprule
				Dataset & \multicolumn{4}{c||}{ORL} & \multicolumn{4}{c||}{MITIndoor-67} & \multicolumn{4}{c}{Scene-15}\\
				\midrule
				Method & NMI & ACC & F-score & Precision & NMI & ACC & F-score & Precision & NMI & ACC & F-score & Precision \\
				\midrule
				SPC$_{\rm{best}}$ & 0.884 & 0.725 & 0.664  & 0.610 & 0.559 & 0.443 & 0.315 & 0.294 & 0.421 & 0.437 & 0.321 & 0.314 \\
				\midrule
				LRR$_{\rm{best}}$ & 0.895  & 0.773 & 0.731 & 0.701 & 0.226 & 0.120 & 0.045 & 0.044 & 0.426 & 0.445 & 0.324 & 0.316\\
				\midrule
				RMSC & 0.854 & 0.704 & 0.623 & 0.582 & 0.342 & 0.232 & 0.123 & 0.121 & 0.564 & 0.507 & 0.437 & 0.425 \\
				\midrule
				AMGL & 0.883 & 0.725 & 0.535 & 0.410 & 0.281 & 0.134 & 0.054 & 0.029 & 0.514 & 0.362 & 0.301 & 0.194\\
				\midrule
				GMC & 0.857 & 0.633 & 0.360 & 0.232 & 0.194 & 0.107 & 0.031 & 0.016 & 0.504 & 0.358 & 0.270 & 0.164\\
				\midrule
				DiMSC & 0.940 & 0.838 & 0.807 & 0.764 & 0.383 & 0.246 & 0.141 & 0.138 & 0.269  & 0.300 & 0.181 & 0.173 \\
				\midrule
				LMSC & 0.921 & 0.819 & 0.762 & 0.713 & 0.484 & 0.368 & 0.237 & 0.228 & 0.531 & 0.497 & 0.401 & 0.364\\
				\midrule
				LT-MSC & 0.930 & 0.795 & 0.768 & 0.766 & 0.546 & 0.431 & 0.290 & 0.279 & 0.571 & 0.574 & 0.465 & 0.452\\
				\midrule
				t-SVD-MSC & \underline{0.993} & \underline{0.970} & \underline{0.968} & \underline{0.946} & 0.750 & 0.684 & 0.562 & 0.543 & 0.858 & 0.812 & 0.788  & 0.743 \\
				\midrule
				ETLMSC & 0.980 & 0.907 & 0.904 & 0.853 & \underline{0.899} & \underline{0.775} & \underline{0.733} & \underline{0.709} & \underline{0.902} & \underline{0.878} & \underline{0.862} & \underline{0.848} \\
				\midrule
				TISRL & \bf{0.995} & \bf{0.975} & \bf{0.976} & \bf{0.961} & \bf{0.925} & \bf{0.886} & \bf{0.845} & \bf{0.824} & \bf{0.908} & \bf{0.888} & \bf{0.875} & \bf{0.864} \\
				\bottomrule
		\end{tabular}}
	\end{table*}

	\subsection{Results and Discussion}
	For each algorithm, Multi-view clustering results of different methods are shown in Table~\ref{Results_all_1}, \ref{Results_all_2}, and \ref{Results_all_3}. In a big picture, the proposed TISRL can achieved the best clustering results in all datasets. 
	
	Comparing to single view clustering, our method has much better clustering results. For example, around 0.265 and 0.253 improvements can be attained by our method comparing to SPC$\rm{_{best}}$ \cite{von2007tutorial} and LRR$\rm{_{best}}$ \cite{von2007tutorial} in the metric of NMI. Actually, for most multi-view clustering algorithms, they achieves better performance than SPC$\rm{_{best}}$ and LRR$\rm{_{best}}$ in most cases, since more information can be utilized for clustering. 
	
	Another interesting observation is that tensor-based multi-view clustering methods, i.e., LT-MSC, t-SVD-MSC, ETLMSC, and TISRL, have better clustering results than others in general. Actually, the best and the second best clustering results of these benchmark datasets are almost obtained by t-SVD-MSC or ETLMSC or TISRL. The reason is that the high-order correlations of multiple views are considered in these methods.
	
	It is notable that the proposed TISRL can get the perfect clustering results in BBCSport, NGs, and MSRCV1 datasets. For the rest datasets, our method also achieves better clustering results than other clustering methods with a remarkable margin. It is reasonable since the rank preserving decomposition accompanied with the t-SVD based low-rank tensor constraint is introduced in the proposed TISRL. Both the specific information and high-order correlations of multi-view data can be investigated fully to improve the clustering performance.
	
	\subsection{Parameters sensitivity and convergence analysis}
\begin{figure*}[!t]
		\centering
		\includegraphics[width=0.7\textwidth]{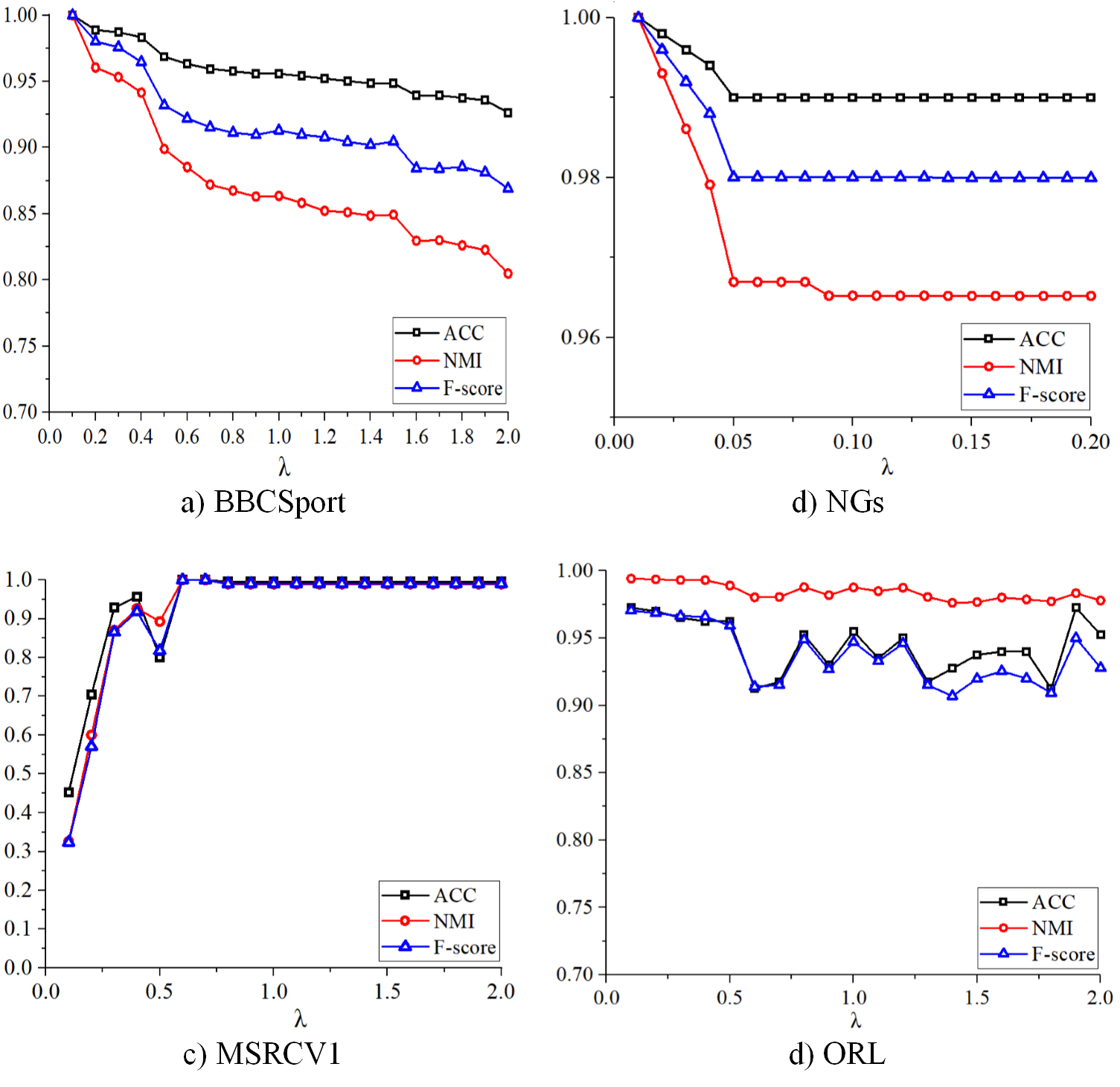}
		\caption{Clustering results of the proposed TISRL on a) BBCSport, b) NGs, c) MSRCV1, and d) ORL with respect to different values of $\lambda$. Clustering results in metrics of NMI, ACC, and F-score are reported here.}
		\label{Para_Tunning}
	\end{figure*}

	A tradeoff parameter, i.e., $\lambda$, is involved in our TISRL. It is worth noting that no extra parameter is added by introducing the rank-preserving decomposition. Clearly, the optimal value of $\lambda$ is decided by dataset's error level. As can be observed in Fig.~\ref{Para_Tunning}, small values of $\lambda$ are preferred by BBCSport and NGs. For MSRCV1 and ORL, clustering results are robust to $\lambda$ on a broad range.
	
\begin{figure*}[!t]
		\centering
		\includegraphics[width=0.7\textwidth]{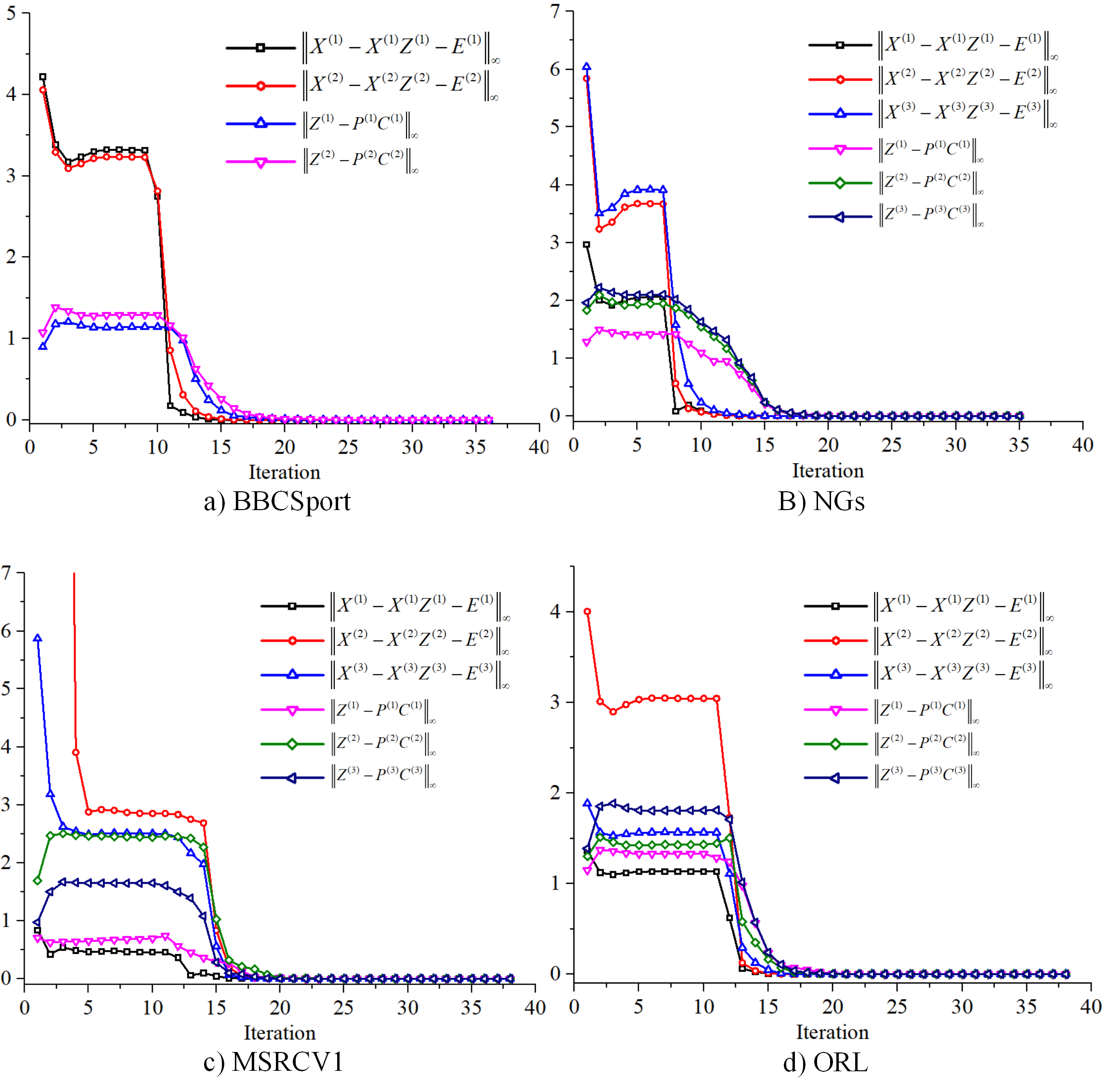}
		\caption{Convergence curves of the proposed TISRL. Experimental results on a) BBCSport, b) NGs, c) MSRCV1, and d) ORL are displayed to illustrate the convergence properties. For MSRCV1, results of the first 3 views are reported.}
		\label{Convergence}
	\end{figure*}

	Regarding to convergence properties, the following curves can be observed:
	\begin{equation}
	\begin{array}{l}
	l_{\rm{error1}}^{(i)}={\left\| {{{\bm{X}}^{(i)}} - {{\bm{X}}^{(i)}}{{\bm{Z}}^{(i)}} - {{\bm{E}}^{(i)}}} \right\|_\infty } ,\\
	l_{\rm{error2}}^{(i)}={\left\| {{{\bm{Z}}^{(i)}} - {{\bm{P}}^{(i)}}{{\bm{C}}^{(i)}}} \right\|_\infty }.
	\end{array}
	\end{equation}
	To be specific, convergence curves on BBCSport, NGs, ORL, and MSRCV1 are reported here. It is notable that the proposed TISRL has similar convergence curves on the rest multi-view datasets. Clearly, Fig.~\ref{Convergence} empirically illustrates that our method can achieve convergence within a small number of iterations.
	
	\section{Conclusions}
	A novel tensor-based intrinsic subspace representation learning for multi-view clustering is proposed in this paper. To fully mine the underlying clustering properties of different views, a rank preserving decomposition is introduced and applied on the original subspace representations of all views, simultaneously, a t-SVD-based low-rank tensor nuclear norm is leveraged to explore the high order correlations among views. Additionally, an augmented Lagrangian multiplier based alternating direction minimization algorithm is also designed. Comparing with ten state-of-the-art clustering methods, experimental results on 9 real-life multi-view datasets verify superiority of the proposed TISRL.
	
	\section*{Acknowledgements}
	
	This work is supported by National Natural Science
	Foundation of China under the Grant No. 61573273, and Fundamental Research Funds for Central Universities under the Grant No. xzy022020050.
	
	\bibliography{wileyNJD-AMA}%
	\bibliographystyle{wileyNJD-AMA}
	
\end{document}